%% file: main.tex
\pdfoutput=1

\documentclass[11pt]{article}

\usepackage[preprint]{acl}

\usepackage{times}
\usepackage{latexsym}
\usepackage{algorithm}
\usepackage[noend]{algpseudocode}
\usepackage{amsmath}
\usepackage{amssymb}
\usepackage{booktabs}
\usepackage{enumerate}
\usepackage{colortbl,xcolor}
\usepackage{multirow}
\usepackage{pgfplots}
\usepackage{caption}
\usepackage{subcaption}    
\usepackage{tcolorbox}
\usepackage{arydshln}

\usepackage{pgfplots}
\pgfplotsset{compat=1.17}
\usepackage{sansmath} 
\usepackage{pgf-pie}
\usepackage{pifont}

\definecolor{soft1}{RGB}{142, 207, 201}  
\definecolor{soft2}{RGB}{255, 190, 122}  
\definecolor{soft3}{RGB}{250, 127, 111}  
\definecolor{soft4}{RGB}{130, 176, 210}  
\definecolor{soft5}{RGB}{174, 199, 232}  
\definecolor{soft6}{RGB}{190, 184, 220}  
\definecolor{soft7}{RGB}{231, 218, 210}  
\definecolor{soft8}{RGB}{153, 153, 153}  
\definecolor{soft9}{RGB}{219, 219, 141}  

\usepackage[T1]{fontenc}

\usepackage[utf8]{inputenc}

\usepackage{microtype}

\usepackage{inconsolata}

\usepackage{graphicx}
\usepackage{caption}
\usepackage{array}
\usepackage{subfiles}
\usepackage{paralist}
\usepackage{enumitem} 
\title{DeepRAG: Thinking to Retrieve Step by Step for Large Language Models}

\author{
 \textbf{Xinyan Guan\textsuperscript{1,2}},
 \textbf{Jiali Zeng\textsuperscript{3}},
 \textbf{Fandong Meng\textsuperscript{3}},
 \textbf{Chunlei Xin\textsuperscript{1,2}},
 \textbf{Yaojie Lu\textsuperscript{1}},
 \\
 \textbf{Hongyu Lin\textsuperscript{1}},
 \textbf{Xianpei Han\textsuperscript{1}},
 \textbf{Le Sun \textsuperscript{1}},
 \textbf{Jie Zhou\textsuperscript{3}}
\\
 \textsuperscript{1}Chinese Information Processing Laboratory, Institute of Software, Chinese Academy of Sciences\\
 \textsuperscript{2}University of Chinese Academy of Sciences\\
 \textsuperscript{3}Pattern Recognition Center, WeChat AI, Tencent Inc, China
\\
\texttt{\{guanxinyan2022,chunlei2021,hongyu,luyaojie,xianpei,sunle\}@iscas.ac.cn} \\
 \texttt{\{lemonzeng,fandongmeng,withtomzhou\}@tencent.com} \\
}

\begin{document}


\maketitle
\begin{abstract}

Large Language Models (LLMs) have shown remarkable reasoning capabilities, while their practical applications are limited by severe factual hallucinations due to limitations in the timeliness, accuracy, and comprehensiveness of their parametric knowledge.
Meanwhile, enhancing retrieval-augmented generation (RAG) with reasoning remains challenging due to ineffective task decomposition and redundant retrieval, which can introduce noise and degrade response quality.
In this paper, we propose DeepRAG, a framework that models retrieval-augmented reasoning as a Markov Decision Process (MDP), enabling reasonable and adaptive retrieval. By iteratively decomposing queries, DeepRAG dynamically determines whether to retrieve external knowledge or rely on parametric reasoning at each step.
Experiments show that DeepRAG improves retrieval efficiency and boosts answer accuracy by 
26.4\%, demonstrating its effectiveness in enhancing retrieval-augmented reasoning.~\footnote{https://github.com/gxy-gxy/DeepRAG}

\end{abstract}

\input{latex/intro}
\input{latex/related}

\input{latex/method}
\input{latex/exp}
\input{latex/analysis}
\input{latex/conclude}
\input{latex/limitation}

\bibliography{custom}
\input{latex/appendix}

\end{document}

%% file: latex/intro.tex
\newcommand{\lemon}[1]{\textcolor{red}{#1}}

\section{Introduction}

Large Language Models (LLMs) have shown considerable promise in reasoning~\cite{plaat2024reasoning}.
Nevertheless, their limitations in capacity and capabilities result in significant issues with factual hallucinations, stemming from challenges related to the timeliness, accuracy, and comprehensiveness of their parametric knowledge~\cite{zhang2023hallucination,huang2023survey}. To mitigate these problems, Retrieval-Augmented Generation (RAG) has been introduced as a promising approach. By incorporating relevant information from knowledge bases or search engines, RAG enhances the factual accuracy of model responses~\cite{zhao2024retrieval}.

\begin{figure}[t]
    \centering
    \includegraphics[width=0.98\linewidth]{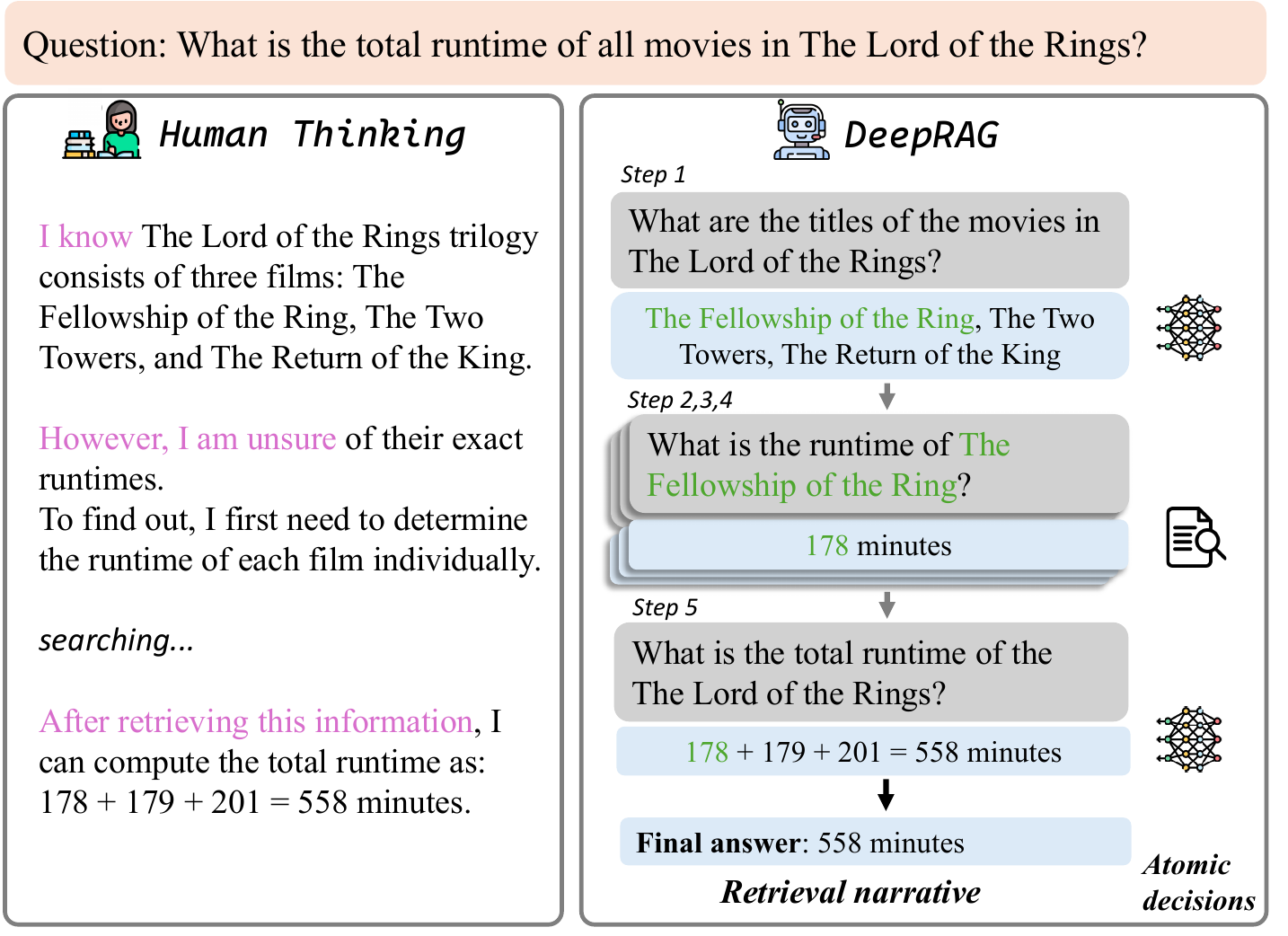}
\caption{Correspondence between human thinking processes and DeepRAG.
Specifically, \textit{retrieval narrative} ensures a structured workflow by generating subqueries that seek additional information based on previous content, and
\textit{atomic decisions} dynamically determines whether to retrieve external knowledge or rely solely on the parametric knowledge for each subquery.}
\vspace{-0.5cm}
    \label{fig:comparison}
\end{figure}

However, enhancing RAG with reasoning still poses several challenges~\cite{gao2025synergizing}. 
One significant issue is that complex queries often necessitate multi-step decomposition to establish a coherent reasoning process~\cite{radhakrishnan2023question,guan2024mitigating}. 
Iterative retrieval has been proposed as a solution to continuously update retrieval results, addressing the dynamic information needs that arise during the generation process~\cite{yue2024inference,wang2025chain}. 
Despite this, LLMs frequently struggle to generate precise and atomic subqueries, which are essential for more effective retrieval and question decomposition~\cite{wu2024divide}.
From the perspective of RAG, iterative retrieval should ideally generate the next atomic query based on the current question and the available information in an adaptive manner.
As illustrated in Figure~\ref{fig:comparison}, the process flows logically from one step to the next.
Specifically, the goal of finding each movie's runtime in \textit{steps 2-4} is derived from \textit{step 1}'s identification of the three titles of the Lord of the Rings series.


Additionally, retrieval is not always essential~\cite{jeong2024adaptive}. Some queries depend on external knowledge (\textit{steps 2-4}), while others can be addressed through the reasoning capabilities of the LLM alone (\textit{step 5} requires summarizing previous information). Moreover, LLMs have shown the ability to function as knowledge bases in their own right~\cite{petroni-etal-2019-language} (such as in \textit{step 1}, where the three movie titles are widely known). 
Unnecessary retrieval can be redundant and may introduce noise, and degrade the quality of generated responses~\cite{chen2023understanding,tan2024blinded,yu2022generate}.

%

To tackle these issues, we introduce \textbf{DeepRAG}, a new framework inspired by how humans search the Internet based on demand. 
This framework aims to enhance reasoning capabilities in retrieval-augmented generation by modeling the process as a Markov Decision Process. DeepRAG incorporates two main components: \textit{retrieval narrative} and \textit{atomic decisions}, which together create a strategic and adaptive retrieval system. As depicted in Figure~\ref{fig:comparison}, the \textit{retrieval narrative} ensures a structured workflow by generating subqueries that seek additional information based on previous content. 
For each subquery, \textit{atomic decisions} dynamically determines whether to retrieve external knowledge or rely solely on the LLM's parametric knowledge.

\begin{figure*}[htbp]
    \centering
          \setlength{\abovecaptionskip}{1pt} 
    \includegraphics[width=0.98\linewidth]{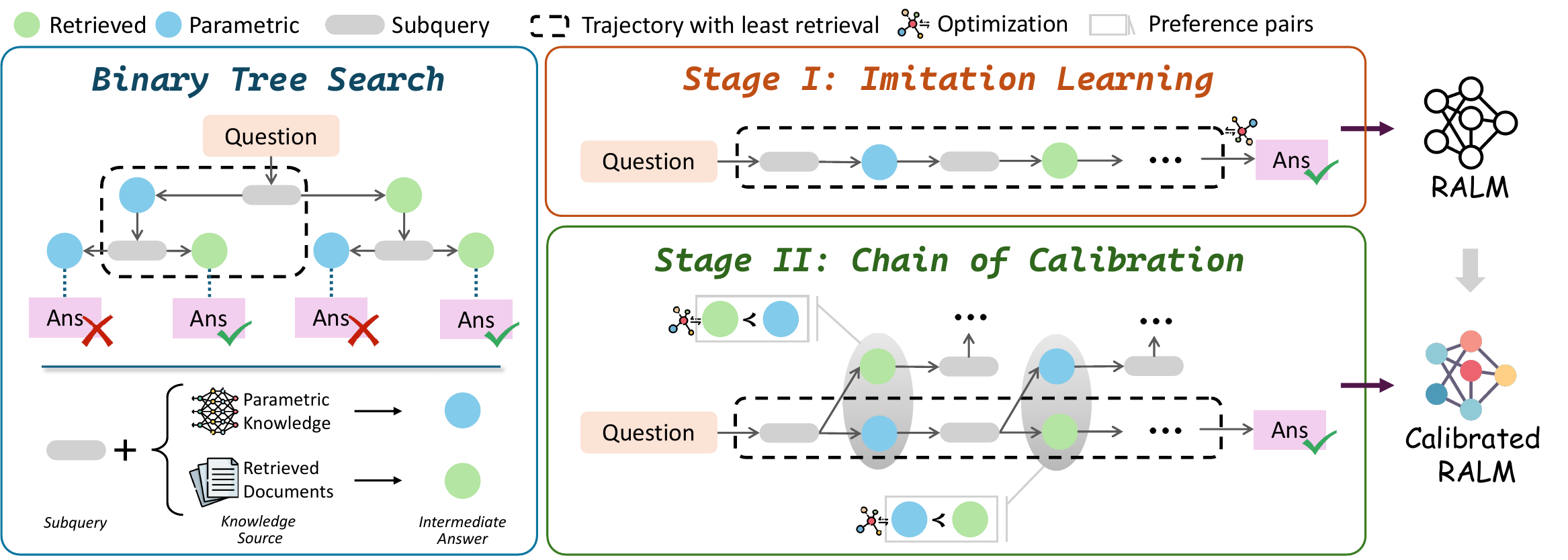}
    \caption{An overview of DeepRAG, our framework comprises three steps: (1) Binary Tree Search, (2) Imitation Learning, and (3) Chain of Calibration.
    Given a set of supervised datasets, we first use binary tree search to synthesize data for imitation learning, allowing the model to learn effective retrieval patterns. Next, we employ binary tree search to generate preference data, further calibrating the LLM's awareness of its knowledge boundaries. }
    \label{fig:main}
\end{figure*}

As illustrated in Figure~\ref{fig:main}, our framework consists of three components:
1) \textbf{Binary Tree Search}, which constructs a binary tree for each subquery related to the given question, exploring paths based on either parametric knowledge or external knowledge.
2) \textbf{Imitation Learning}, which extracts the reasoning process that leads to the correct final answer with minimal retrieval cost based on Binary Tree Search, enabling the model to learn the pattern of ``subquery generation – \textit{atomic decision} – intermediate answer''.
3) \textbf{Chain of Calibration}, which calibrates the LLM's internal knowledge by refining each atomic decision, enabling it to make accurate \textit{atomic decisions} about the necessity of retrieval.
By explicitly enhancing the LLM's ability to recognize its own knowledge limits, we can train any model in an end-to-end manner, allowing it to dynamically decide when and what to retrieve.


We validate the effectiveness of DeepRAG across in-distribution, out-of-distribution, time-sensitive, and heterogeneous knowledge base datasets.
Experimental results show that DeepRAG significantly outperforms existing methods, achieving a 21.99\% increase in accuracy while also enhancing retrieval efficiency. Further analysis indicates that DeepRAG demonstrates a stronger correlation between its retrieval decisions and parametric knowledge, suggesting more effective calibration of knowledge boundaries.

%% file: latex/related.tex
\section{Related Work}

\paragraph{Adaptive Retrieval-Augmented Generation}
%
%
%
%



Existing adaptive RAG approaches can be broadly categorized into three types: classifier-based methods \cite{cheng2024unified,jeong2024adaptive} requiring additional linear head training for retrieval decisions, confidence-based methods \cite{jiang2023flare,su-etal-2024-dragin,dhole2025retrieveretrieveuncertaintydetection} relying heavily on threshold-dependent uncertainty metrics, and LLM-based methods~\cite{asai2023self,zhang2024retrievalqa} generating retrieval decisions but often fail to accurately recognize their knowledge boundaries, making it unreliable to delegate retrieval timing decisions to the model.
Our method leverages the inherent generative capabilities of LLMs to explore knowledge boundaries in RAG settings. This design maintains the model's native generation abilities while eliminating the need for additional parameters or unreliable uncertainty metrics.



\paragraph{Reasoning in Retrieval-Augmented Generation}
\label{relate-work:reasoning}

Recent advances in RAG have increasingly emphasized the integration of reasoning capabilities. 
Search-o1~\cite{li2025search} incorporates retrieval into inference to build an agentic system, while its application is limited to reasoning models~\cite{chen2025towards}.
Self-RAG~\cite{asai2023self} and Auto-RAG~\cite{yu2024autorag} enhance reasoning through automatic data synthesis within retrieval-augmented frameworks, while AirRAG~\cite{feng2025airragactivatingintrinsicreasoning} combines Monte Carlo Tree Search with self-consistency techniques.
These methods, however, often depend heavily on extensive retrieval or sampling overhead.
More recent developments have explored reinforcement learning to enhance retrieval quality~\cite{jin2025search,song2025r1,gao2024smartrag}, while these methods generally overlook retrieval efficiency in their reward function. 
In contrast, DeepRAG offers a flexible, end-to-end solution that enables arbitrary models to retrieve information step by step as needed, based on their evolving reasoning process.

\paragraph{Knowledge Boundary}

%
%
LLMs struggle to accurately distinguish between what they know and what they don't know~\cite{yin2023large,kapoor2024large,yin2024benchmarking}.
Additional fine-tuning~\cite{kapoor-etal-2024-calibration} or precise probing~\cite{cheng2024unified} is typically required to calibrate the model's cognition. 
Our approach explores knowledge boundaries in RAG settings.

%% file: latex/method.tex
\section{Thinking to Retrieve Step by Step}


In this section, we present our proposed method, DeepRAG. At its core, DeepRAG models the process of question decomposition, atomic decisions, and final answer generation as a Markov Decision Process. 
%
%
Given a set of supervised datasets, we first use binary tree search to synthesize data for imitation learning, allowing the model to learn effective retrieval patterns. Next, we employ binary tree search to generate preference data, further calibrating the LLM's awareness of its knowledge boundaries. 
In the following subsections, we provide a detailed description of each component of DeepRAG.

%

\subsection{Overview of the MDP Modeling}
\label{overview}

We formalize the step-by-step reasoning process of retrieval-augmented generation as a Markov Decision Process~\cite{sutton2018reinforcement}, represented by the tuple \((\mathcal{S}, \mathcal{A}, P, R)\), where \(\mathcal{S}\) denotes the set of states, \(\mathcal{A}\) represents the set of actions, \(P\) defines the transition dynamics, and \(R\) specifies the reward function.

\paragraph{States.}
At each step \(t\), the state \(s_t \in \mathcal{S}\) represents the partial solution to the original question. We denote
$s_t = \bigl[x,\,(q_1, r_1),\;\dots,\,(q_{t}, r_{t})\bigr]$,
where \(x\) is the input question, \(q_i\) refers to the  \(i\)-th subquery, and \(r_i\) refers to the  \(i\)-th intermediate answer (and any retrieved documents based on \(q_i\)).

\paragraph{Actions.}
At state \(s_t\), the model selects an action \(a_{t+1} = (\sigma_{t+1}, \delta_{t+1}) \in \mathcal{A}\), which consists of two sub-decisions:

1. \emph{Termination decision}:  Given the partial solution \(s_t\), the model makes a binary decision \(\sigma_{t+1} \in \{\texttt{continue}, \texttt{terminate}\}\) to determine whether to proceed with generating the next subquery \(q_{t+1}\) or finalize the answer \(o\). 

2. \emph{Atomic decision}: For each subquery \(q_{t+1}\), the model decides whether to retrieve external knowledge or rely solely on its parametric knowledge. Formally, this decision is represented as \(\delta_{t+1} \in \{\texttt{retrieve}, \texttt{parametric}\}\).

\paragraph{Transitions.}  
After executing the action \(a_{t+1} = (\sigma_{t+1}, \delta_{t+1})\) in state \(s_t\), the environment updates the state to \(s_{t+1}\) based on transition dynamics $P$.  

Specifically, if \(\sigma_{t+1} = \texttt{terminate}\), the process concludes by generating the final answer \(o\), resulting in the terminal state \(s_{t+1} = \bigl[x,\,(q_1,r_1),\;\dots,\,(q_{t}, r_{t}), o\bigr]\). Otherwise, it generates the next subquery \(q_{t+1}\).  

If \(\delta_{t+1} = \texttt{retrieve}\), the model retrieves documents \(d_{t+1}\) and generates an intermediate answer \(ia_{t+1}\) for subquery \(q_{t+1}\). Otherwise, it relies on parametric knowledge to generate the intermediate answer. 
The response \(r_{t+1}\) is set as \([d_{t+1}, ia_{t+1}]\) (if retrieved) or \(ia_{t+1}\) (if not). The updated state is  
\(s_{t+1} = \bigl[x,\,(q_1,r_1),\;\dots,\,(q_{t+1}, r_{t+1})\bigr]\).

\paragraph{Rewards.}  
The reward function evaluates the state based on answer correctness and retrieval cost, applied only after generating the final answer \(o\). Formally,  
$R\bigl(s_{t+1} = s_t+[o]\bigr) = -C(o) \times T(s_t)$,
where \(C(o)\) indicates correctness (\(1\) if correct, \(\infty\) otherwise), and \(T(s_t)\) represents the total retrieval cost in state \(s_t\).
Therefore, this reward prioritizes answer correctness while encouraging the model to reduce retrieval cost as much as possible. \label{line:783-785}

\subsection{Binary Tree Search}


Building on this formulation, LLM iteratively decomposes a given question into subqueries, each derived from previously acquired information. 
The detailed generation instruction is outlined in Appendix~\ref{template}, with the answer format below.  
\begin{figure}[h]
\begin{tcolorbox}[colframe=cyan!40!black,title=\textbf{Answer format}]
\footnotesize
\textbf{Question}: <Question>

\textbf{Follow up}: <Subquery1>

Let's search the question in Wikipedia.

Context: <Paragraph Text>

\textbf{Intermediate answer}: <Intremediate Answer1>

\textbf{Follow up}: <Subquery2>

\textbf{Intermediate answer}: <Intermediate Answer2>

......

\textbf{So the final answer is}: <Answer>
\end{tcolorbox}
\end{figure}
\vspace{-0.3cm}


Then, we implement a binary tree search to construct reasoning paths that integrate different retrieval strategies for each subquery. As illustrated in Figure~\ref{fig:main}, given a question, the model generates the \(i\)-th subquery and explores two answering strategies: directly leveraging parametric knowledge (blue node) or retrieving external documents (green node). 
Therefore, we can construct a binary tree for each subquery related to the given question, exploring paths based on either parametric knowledge or external knowledge.

\subsection{Imitation Learning}
We present an algorithm that leverages binary trees to identify the optimal reasoning process that leads to the correct final answer while minimizing retrieval costs, corresponding to the highest reward as defined in Section~\ref{overview}.
Based on the synthesized optimal reasoning data, we fine-tune the model to improve its termination and atomic decisions while enhancing its query decomposition capabilities and generating faithful intermediate answers.

\paragraph{Synthesizing Data}  As shown in Alg.~\ref{algorithm:imi}, we employ a priority queue to maintain reasoning trajectories based on their retrieval costs. This allows us to efficiently explore potential reasoning paths by iteratively constructing and evaluating them until either finding a correct answer or exhausting all viable options within specified constraints. 
For instances where no correct answer can be obtained after exhausting all options, we discard them.

Through the synthesis process above, the training dataset obtained contains an adaptive reasoning process, which can be used to facilitate arbitrary LLMs in enhancing the RAG capabilities.
\input{latex/imi_algorithm}
\vspace{-0.2cm}
\paragraph{Training Objective} We implement a masked loss function for the retrieved documents to prevent the model from learning irrelevant or noisy text that could negatively impact its performance. The detailed objective is shown in Appendix~\ref{eq1}.
\vspace{-0.2cm}
\subsection{Chain of Calibration}
Building on the markov process in Section~\ref{overview}, we identify four key optimization aspects for DeepRAG: termination and atomic decisions, query decomposition, and intermediate answer generation. Unlike the others, \textit{atomic decisions} require the model to recognize its own knowledge boundaries to make precise judgments.  


We propose a method that dynamically optimizes atomic decisions for each subquery, rather than training LLMs on complete reasoning paths. 
Our approach consists of two key components: (1) synthesizing preference data to determine when retrieval is necessary, and (2) fine-tuning the LLM with this data using Chain of Calibration training to enhance its ability to make informed atomic decisions based on its internal knowledge boundaries.

%



%
\vspace{-0.1cm}
\paragraph{Synthesizing Preference Data}
First, we identify an optimal path with minimal retrieval based on Alg.~\ref{algorithm:imi} using the model trained in Stage I. This provides the optimal atomic decision for each subquery, determining whether retrieval is necessary.  
From this path, we construct preference pairs for each subquery to indicate the preferred retrieval choice. For example, in Figure~\ref{fig:main}, the optimal path may suggest answering the first subquery using parametric knowledge while requiring document retrieval for the second. 
Accordingly, we generate preference pairs favoring parametric knowledge for the first subquery and retrieval for the second.
%
This process enables LLMs to learn when to retrieve external information, thereby improving its ability to maximize the use of parametric knowledge and reducing unnecessary retrievals.

%

%

\vspace{-0.1cm}
\paragraph{Chain of Calibration Objective}

We fine-tune the LLM using a Chain of Calibration objective on our synthesized preference data. 
Given the $i$-th subquery and a state $s_i = [x,(q_1, r_1), \cdots, (q_{i-1}, r_{i-1})]$, we have two distince intermediate answer $r_i^1 = a_i^1$ and $r_i^2 = (d_i, a_i^2)$. Based on the synthesis process above, we can tag which $r_i$ is preferred and optimize it. The detailed equation is shown in Appendix~\ref{eq2}.

%% file: latex/imi_algorithm.tex
\begin{algorithm}
\footnotesize
\caption{Data Construction for Stage I}
\label{algorithm:imi}
\begin{algorithmic}[1]
\Require Question $x$, answer $y$, language model $\mathcal{M}$, Retriever $\mathcal{R}$, max history length $T$
\Ensure Optimal reasoning process $s^*$ or $null$
\State Initialize priority queue $\mathcal{PQ} \gets \{([x], 0)\}$ 

\Comment{(trajectory, retrieval count)}
\While{$\mathcal{PQ}$ is not empty}
    \State $(h, r) \gets \mathcal{PQ}.dequeue()$ 
    
    \Comment{Get trajectory with lowest retrieval count}
    \State $q \gets \mathcal{M}(h)$ \Comment{Subquery Generation}
    \If{$ShouldAnswer(q)$ or $length(h) > T$}
        \State $o \gets \mathcal{M}(h, q)$ \Comment{Final answer}
        \If{$IsEqual(o, y)$}
            \Return $h$
        \EndIf
    \Else
        \State $a \gets \mathcal{M}(h, q)$ \Comment{Direct answer}
        \State $\mathcal{PQ}.enqueue(([h, (q, a)], r))$
        \State $d \gets \mathcal{R}(q)$ \Comment{Retrieve document}
        \State $a \gets \mathcal{M}(h, q, d)$ \Comment{Retrieved answer}
        \State $\mathcal{PQ}.enqueue(([h, (q, (d, a))], r+1))$
    \EndIf
\EndWhile
\State \Return $null$
\end{algorithmic}
\end{algorithm}

%% file: latex/exp.tex
\vspace{-0.2cm}
\section{Experiment}
\vspace{-0.1cm}
\subsection{Datasets}
We use six open-domain QA datasets for our experiments. 
We treat training datasets as \textit{in-distribution}, and unseen ones as \textit{out-of-distribution}.
The in-distribution datasets include HotpotQA~\cite{yang2018hotpotqa}, and 2WikMultihopQA~\cite{ho-etal-2020-constructing}, and the out-of-distribution datasets consist of CAG~\cite{pan2024not}, PopQA~\cite{mallen2022not}, WebQuestions~\cite{berant2013semantic}, and MuSiQue~\cite{trivedi2022musique}.
Specifically, we employ the time-sensitive subset of CAG to evaluate temporal reasoning capabilities.
Furthermore, WebQuestions is built upon Freebase to assess model robustness when information may be absent from the knowledge base.

\vspace{-0.1cm}
\subsection{Baselines}
We use the following baselines to evaluate the performance: 
\textbf{CoT}~\cite{wei2022chain} and \textbf{CoT*}, which employ 8-shot examples extracted from the training dataset. The asterisk (*) indicates the model output was trained using the same data employed for training the DeepRAG.
\textbf{CoT-Retrieve} and \textbf{CoT-Retrieve*} augment the eight examples in the context with retrieved relevant documents based on the query.
\textbf{IterDRAG}~\cite{yue2024inference} refers to decomposing the question and answer step by step based on in-context learning.
\textbf{AutoRAG}~\cite{yu2024autorag} uses trained models to iteratively decompose questions and retrieve relevant documents for answering.
\textbf{Search-o1}~\cite{li2025search} leverages special tokens to prompt reasoning models to autonomously invoke retrieval as needed.
\textbf{UAR}~\cite{cheng2024unified} employs a trained classifier to determine when to retrieve.
\textbf{FLARE}~\cite{jiang2023flare} and \textbf{DRAGIN}~\cite{su-etal-2024-dragin} are confidence-based method that decide the timing of retrieval based on token importance and uncertainty.
\textbf{TAARE}~\cite{zhang2024retrievalqa} allows the LLM itself to determine when retrieval is needed.

\input{table/main-exp}
\vspace{-0.1cm}
\subsection{Implementation Details}



We train our target model on two QA datasets: HotpotQA and 2WikiMultihopQA. For imitation learning, we randomly sample 4,000 examples from each dataset. 
To enhance the model's question decomposition and context-based generation capabilities, we employ Qwen-2.5-72B to generate the gray (query decomposition) and green nodes (retrieved answers) in Figure~\ref{fig:main}, and use the target model to generate the blue nodes (parametric answers) for data synthesis.
For chain of calibration, we sample an additional 1,000 examples from each dataset.
The performance is evaluated using Exact Match (EM) and F1 score.

Following \citet{su-etal-2024-dragin}, we adopt BM25 for retrieval 
and Wikipedia\footnote{\url{https://dl.fbaipublicfiles.com/dpr/wikipedia_split/psgs_w100.tsv.gz}} as knowledge base.
For time-sensitive questions in CAG, we utilize the dataset-provided up-to-date passages as knowledge base.
We selected Llama-3-8B-Instruct~\cite{dubey2024llama}, Qwen-2.5-7B and Qwen-2.5-32B~\cite{qwen2}  as our target model. 
To implement Search-o1, we employ the distillation series of DeepSeek-R1~\cite{guo2025deepseek}, as the method depends on reasoning models.

\vspace{-0.1cm}
\subsection{Overall Results}
The results in Table~\ref{tab:main-exp} demonstrate DeepRAG's superior performance and robustness across different scenarios.


\textbf{DeepRAG demonstrates superior performance across most datasets via thinking to retrieve step by step.} 
Our method consistently outperforms existing approaches across various backbones and model sizes.
Compared to reasoning-based and adaptive RAG baselines, DeepRAG outperforms across all datasets, demonstrating the effectiveness of the structured \textit{retrieval narrative} and on-demand  \textit{atomic decisions}.
Specifically, the limited performance of IterDRAG highlights the necessity of learning both query decomposition and faithful answering.
Confidence-based methods like FLARE struggle to determine the optimal retrieval timing due to their reliance on unstable, predefined metrics. 
Moreover, we observe that confidence-based methods suffer from instability, as their performance is highly sensitive to threshold selection. 
Meanwhile, iterative retrieval methods like Auto-RAG often fall into continuous retrieval loops when no highly relevant information is found.

\vspace{-0.1em}
\textbf{DeepRAG exhibits remarkable generalization capabilities and robustness in time-sensitive and out-of-distribution settings.}
In the time-sensitive dataset CAG, DeepRAG performs well compared to other adaptive and reasoning retrieval methods. 
It is worth noting that CoT-Retrieve outperforms it on CAG. 
We attribute this to the core challenge of the time-sensitive setting is to trigger retrieval most of the time.
%
Furthermore, DeepRAG achieves substantial F1 score improvements of 2.63 and 4.57 on PopQA and WebQuestions respectively, even in scenarios where relevant information may be sparse or missing from the knowledge base.

\vspace{-0.05cm}
\textbf{By learning from self-synthesized data, DeepRAG effectively explores knowledge boundaries while minimizing hallucination risks.} 
TAARE often underperforms direct retrieval methods, highlighting the mismatch between its internal knowledge and verbose.
Moreover, aggressive fine-tuning approaches like CoT* and CoT-Retrieve* degrade model performance by forcing the model to learn knowledge beyond its knowledge boundaries.
DeepRAG carefully preserves model capabilities during fine-tuning by leveraging self-synthesized data, effectively preventing additional hallucination while maintaining performance.

%% file: table/main-exp.tex
\begin{table*}[t]
  \centering
      \setlength{\abovecaptionskip}{0.5pt} 

  \resizebox{\linewidth}{!}{
    \begin{tabular}{cccccccccccccccc}
    \toprule
           \multirow{3}[3]{*}{\textbf{Types}} & \multirow{3}[3]{*}{\textbf{Methods}} & \multicolumn{4}{c}{\textit{in-distribution}} &       & \multicolumn{8}{c}{\textit{out-of-distribution}} & \multirow{3}[3]{*}{\textbf{Avg}} \\
\cmidrule{3-6}\cmidrule{8-15}          & \multicolumn{1}{c}{} & \multicolumn{2}{c}{Hotpot QA} & \multicolumn{2}{c}{2WikiMultihopQA} &       & \multicolumn{2}{c}{CAG} & \multicolumn{2}{c}{PopQA} & \multicolumn{2}{c}{Web Question} &  \multicolumn{2}{c}{MuSiQue} &  \\
     &  & EM    & F1    & EM    & F1    &       & EM    & F1    & EM    & F1    & EM    & F1    & EM    & F1    &  \\
    \midrule
    \rowcolor{gray!20}
    \multicolumn{16}{c}{\textit{Llama-3-8B}} \\
    \multirow{6}[2]{*}{Reasoning} & CoT   & 27.20  & 37.75  & 28.20  & 34.85  &       & 7.17  & 10.41  & 21.20  & 25.33  & 25.20  & 40.56  & 13.70  & 22.97  & 24.54   \\
          & CoT-Retrieve & 34.90  & \underline{46.85}  & 35.80  & 43.41  &       & \textbf{55.45}  & \textbf{64.08}  & 32.80  & 45.87  & 22.90  & 39.22  & 19.10  & 28.18  & 39.05  \\
          & CoT*  & 21.80  & 31.69  & 25.60  & 30.89  &       & 5.30  & 7.58  & 23.10  & 25.31  & 26.80  & 40.20  & 4.80  & 13.85  & 21.41  \\
          & CoT-Retrieve* & 22.50  & 32.15  & 23.70  & 29.21  &       & 44.86  & 55.69  & 38.70  & 45.64  & 17.60  & 29.20  & 5.70  & 11.60  & 29.71  \\
        & IterDRAG & 23.20  & 30.95  & 19.60  & 24.80  &       & 38.32  & 46.18  & 22.70  & 34.53  & 15.90  & 26.79  & 12.40  & 17.75  & 26.09 \\
        & Auto-RAG & 25.80  & 36.09  & 23.00  & 30.09  &       & 49.22  & 59.61  & 27.80  & 42.02  & 17.40  & 32.94  &  19.10  & 28.33  & 32.62 \\
        & Search-o1 & 14.80  & 24.08  & 22.20  & 27.10  &      & 3.43  & 6.61  & 10.30  & 13.54  & 15.30  & 29.60  & 5.40  & 11.98  & 15.36  \\

    \hdashline
    \multirow{4}[2]{*}{Adaptive} & FLARE & 23.80  & 32.88  & 30.30  & 37.45  &       & 34.89  & 43.45  & 28.80  & 40.61  & 28.80  & 40.61  & 14.50  & 23.57  & 31.64   \\
          & DRAGIN & 27.60  & 38.05  & 29.10  & 35.68  &       & 4.05  & 7.18  & 22.60  & 28.53  & 21.20  & 38.72  & 11.80  & 19.97  & 23.71  \\
          & UAR & 29.70 &	40.66 &	34.80 &	42.40 	&&	\underline{52.96} &	61.53 &	33.00 &	45.95 &	22.70 &	39.10 &	19.10  & 28.38  & 37.52 \\
          & TAARE & 30.60  & 41.43  & 35.20  & 42.85  &       & \underline{52.96}  & 61.59  & 33.20  & 46.01  & 23.40  & 39.56  & 18.60  & 27.55  & 37.75 \\
    \hdashline
    \multirow{2}[2]{*}{Ours} & DeepRAG-Imi & \underline{35.10}  & 46.59  & \underline{47.20}  & \underline{52.33}  &       & 50.47  & 59.55  & \textbf{43.60}  & \textbf{48.50}  & \underline{30.00}  & \underline{41.76}  & \underline{22.30}  & \textbf{30.46}  & \underline{42.32}   \\
          & DeepRAG  &  \textbf{40.70}  & \textbf{51.54}   & \textbf{48.10}  & \textbf{53.25}  &       & \underline{52.96}  & \underline{61.92}  & \underline{42.50}  & \underline{47.80}  & \textbf{32.70}  & \textbf{45.24}  &  \textbf{22.50}  & \underline{30.40}  & \textbf{44.13}\\
    \midrule
    \rowcolor{gray!20}
    \multicolumn{16}{c}{\textit{Qwen-2.5-7B}} \\
    \multirow{5}[2]{*}{Resaoning} & CoT   & 18.90  & 27.81  & 23.40  & 28.97  &       & 3.12  & 5.71  & 15.20  & 19.20  & 18.30  & 34.86  & 5.60  & 13.12  & 17.85   \\
          & CoT-Retreive & 24.90  & 34.78  & 18.60  & 23.44  &       & 41.43  & 51.47  & 27.30  & \underline{41.20}  & 15.10  & 29.84  & 6.70  & 15.10  & 27.49   \\
          & CoT*  & 17.60  & 26.15  & 25.10  & 29.62  &       & 3.12  & 5.62  & 7.90  & 11.06  & 15.60  & 32.45  & 4.70  & 13.40  & 16.03   \\
          & CoT-Retrieve* & 23.40  & 32.29  & 22.40  & 27.51  &       & 43.30  & 54.51  & 26.60  & 35.46  & 13.80  & 25.60  & 6.20  & 12.85  & 26.99 \\
        & IterDRAG & 13.70  & 26.84  & 9.30  & 20.47  &       & 21.81  & 39.59  & 18.00  & 31.44  & 12.50  & 26.95  &  9.20  & 17.25  & 20.59 \\
        & Search-o1 & 11.60  & 16.95  & 22.00  & 25.02  &      & 3.43  & 4.78  & 4.40  & 7.61  & 7.70  & 19.97  & 2.10  & 7.48  & 11.09  \\
    \hdashline
    \multirow{4}[2]{*}{Adaptive} & FLARE & 23.40  & 32.06  & 21.80  & 26.51  &       & 34.89  & 42.62  & 19.00  & 28.24  & 16.10  & 31.89  &  8.40  & 15.15  & 25.00  \\
          & DRAGIN & 16.70  & 24.60  & 12.40  & 16.76  &       & 3.43  & 5.45  & 12.00  & 15.80  & 17.40  & 32.43   & 4.20  & 7.98  & 14.10  \\
          & UAR   & 24.50  & 34.22  & 23.90  & 28.20  &       & 34.89  & 43.92  & 27.00  & 40.47  & 16.60  & 32.28  & 7.10  & 15.62  & 27.39  \\
          & TAARE & 25.30  & 35.03  & 21.30  & 25.67  &       & 40.81  & 50.78  & 27.00  & 40.92  & 18.20  & 33.14  & 6.90  & 15.46  & 28.38  \\
    \hdashline
    \multirow{2}[2]{*}{Ours} & DeepRAG-Imi & \underline{30.40}  &\underline{ 39.44}  & \underline{32.00}  &\underline{38.32} &       & \underline{47.98}  & \underline{56.99} & \underline{37.50}  & 40.72  & \underline{23.90}  & \underline{38.62}  & \underline{16.50}  & \underline{24.67}  & \underline{35.59}  \\
          & DeepRAG  & \textbf{32.10}  & \textbf{41.14}  & \textbf{40.40}  & \textbf{44.87}  &       & \textbf{51.09}  & \textbf{59.76}  & \textbf{40.60}  & \textbf{43.19}  & \textbf{24.20}  & \textbf{38.83}  & \textbf{19.50}  & \textbf{32.35}  & \textbf{39.00 } \\

    \midrule 
    \rowcolor{gray!20}
    \multicolumn{16}{c}{\textit{Qwen-2.5-32B}} \\
    
    \multirow{4}[2]{*}{Reasoning} & CoT   & 20.6  & 30.62  & 24.4  & 30.94  & & 3.12  & 5.42  & 10.9  & 14.45 & 9.7   & 26    & 9.5   & 18.26 & 16.99  \\
          & CoT-Retrieve & 28.6  & 39.43  & 27.9  & 36.73  & & 39.56 & 49.97  & 33.8  & 45.91 & 17.2  & 34.15 & 12.9  & 21.98 & 32.34  \\
          & Iter-DRAG  & 22.9  & 38.26  & 19.6  & 35.70  & & 33.02 & 45.61 & 20.3  & 33.2  & 13.3  & 27.57 & 17.6  & \textbf{27.8}  & 27.91  \\
          & Search-o1 & 34    & 45.64  & 29.1  & 35.12  &  & 19    & 24.35  & 23.1  & 30.69 & 17.9  & 35.11 & 16.4  & 25.6  & 28.00  \\
        \hdashline
    Ours & DeepRAG & \textbf{36.2}  & \textbf{46.90}  & \textbf{46.3}  & \textbf{50.50}  & & \textbf{52.02} & \textbf{61.42} & \textbf{46.3}  & \textbf{49.2}  & \textbf{28.1} & \textbf{43.27}& \textbf{19.60}  & 27.47  & \textbf{42.27}  \\

    \bottomrule
    \end{tabular}%
    }
    \caption{The overall experimental results of DeepRAG and other baselines on five benchmarks. The best/second best scores in each dataset are
\textbf{bolded}/\underline{underlined}. DeepRAG-Imi (Stage I) and DeepRAG (Stage II) both demonstrate superior performance compared to existing methods across all test scenarios.}
\vspace{-0.2cm}
  \label{tab:main-exp}%
\end{table*}%




%% file: latex/analysis.tex
\section{Analysis}
\subsection{Retrieval Efficiency}




To evaluate the efficiency of our method, we compare the average number of retrievals on the WebQuestions dataset and report the average computation time per query. The computation time is measured on an H20*8 machine.
As shown in Table~\ref{tab:efficiency}, We have the following observations:
1) DeepRAG can achieve higher accuracy with relatively lower retrieval costs, attributed to its dynamic usage of internal knowledge.
2) Confidence-based approaches demonstrate limited robustness across datasets.  
For instance, neither FLARE nor DRAGIN triggers retrieval under the default confidence threshold in the WebQuestions dataset.
%
3) Iterative retrieval-based methods typically require numerous retrieval operations.
Therefore, efficient adaptive retrieval methods like DeepRAG become crucial for optimizing resource utilization while maintaining performance.

\input{table/efficiency}
\subsection{Relevance to Parametric Knowledge}

In this section, we investigate the relationship between retrieval needs and internal knowledge to demonstrate how effectively \textit{atomic decisions} explores the knowledge boundary.

Ideally, models should initiate retrieval for queries beyond their parametric knowledge while utilizing their existing knowledge for familiar queries. We use CoT results as an indicator of whether the model can answer questions using its parametric knowledge. Then, we analyze whether other adaptive retrieval methods align with this pattern of parametric knowledge utilization.

We report four metrics. 
F1 score and Accuracy serve as basic performance measures, while balanced accuracy and Matthews Correlation Coefficient(MCC)~\cite{Wikipedia_PhiCoefficient} are employed to account for the class imbalance between retrieval-required and retrieval-not-required cases.

As shown in Table~\ref{tab:relevance}, we find that: 1) 
DeepRAG demonstrates superior relevance performance across F1, balanced accuracy, and MCC metrics.
This suggests that DeepRAG successfully identifies retrieval necessity by exploring knowledge boundary;
2) While FLARE, DRAGIN, and TAARE exhibit high accuracy scores, their relatively low balanced accuracy and MCC scores suggest they mainly succeed in retrieval-required cases but struggle to properly avoid unnecessary retrievals.

\input{table/relevance}

\begin{figure}[t]
    \setlength{\abovecaptionskip}{0.5pt} 
    \centering
    \includegraphics[width=0.95\linewidth]{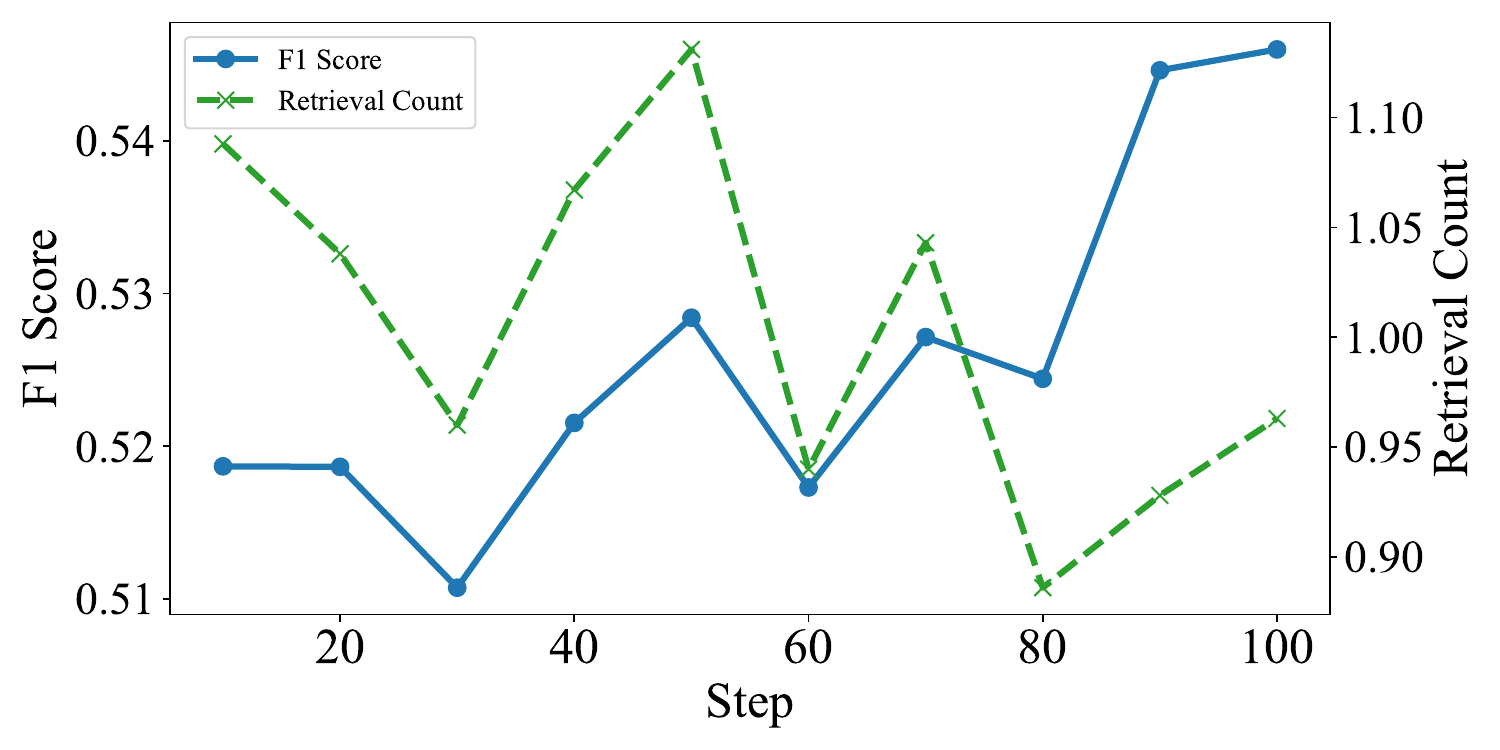}
    \caption{Experiment result and retrieval efficiency on 2WikiMultihopQA under RL setting.}
    \label{fig:rl-fig}
\end{figure}
\input{latex/fig_ablation}

\input{figure/pie}
\subsection{Effectiveness under RL Setting}
\label{rl}
Recently, reinforcement learning (RL) has demonstrated remarkable success in enhancing model capabilities across various domains~\cite{liu2025auto, wen2024policy,guan2024search}. Building upon this, we further explore the potential of DeepRAG by incorporating reinforcement learning. Specifically, we initialize from DeepRAG-Imi and optimize Stage II using the GRPO objective~\cite{shao2024deepseekmath}. The detailed implementation can be found in Appendix~\ref{appendix:rl}, and our code is released in GitHub repository.~\footnote{https://github.com/gxy-gxy/Search-R1-for-DeepRAG/tree/main}

Figure~\ref{fig:rl-fig} presents the training dynamics of our RL-enhanced model. The results reveal an encouraging trend: as training progresses, the F1 score on 2WikiMultihopQA gradually improves while the average number of retrievals decreases. This demonstrates that our reward design effectively guides the model to achieve better performance with more efficient retrieval behavior.


\subsection{Question Decomposition Effectiveness}



We systematically analyze the effectiveness of question decomposition in \textit{retrieval narrative}. 
As shown in Figure~\ref{pie}, we present the distribution of subquery counts and retrieval attempts for different questions. Most questions require 3-5 decomposition steps, while retrieval attempts are primarily concentrated within 0-2 rounds. This demonstrates that DeepRAG effectively decomposes questions while minimizing redundant retrieval.

Moreover, we analyze the average counts of WH-words, nouns, verbs, and conjunctions in subqueries, as shown in Figure~\ref{fig:atomic}. DeepRAG decomposes atomic queries with fewer pronouns and conjunctions, indicating its concise and effective query decomposition strategy.

\subsection{Different Inference Strategy}

To gain a deep insight into the effectiveness of \textit{atomic decision}, we evaluate DeepRAG's performance under two extreme scenarios: relying solely on internal knowledge (retrieve only) and using retrieval in each subquery (parametric only).
As shown in Figure~\ref{fig:analysis1}, parametric only yields poor performance, while retrieve only achieves relatively higher accuracy but incurs substantial retrieval costs.
DeepRAG achieves superior performance by adaptively selecting between internal and external knowledge sources.
\input{table/word_count}

Moreover, DeepRAG outperforms the retrieve only approach because retrieval can hinder model
performance due to long context or \hspace{-0.2em} irrelevant \hspace{-0.1em}knowledge in\hspace{-0.05em} certain \hspace{-0.1em}scenarios.

\vspace{-0.1cm}
\subsection{Ablation Study}
In this section, we conducted experiments to validate the effectiveness of DeepRAG's data construction and training process. 

%


For Imitation Learning, we compare our default strategy of selecting paths with minimal retrieval cost against two alternative approaches: maximum retrieval cost (\textit{most}) and random path selection (\textit{random}).
As shown in Figure~\ref{fig:ablation1}, DeepRAG-Imi achieves lower retrieval costs and higher average performance compared to both the \textit{most} and \textit{random} methods.


For Chain of Calibration, we compare our default approach of constructing preferences based on nodes from optimal paths against two alternatives: constructing pairs for all nodes and constructing sentence-level partial order pairs based on retrieval efficiency.
As shown in Figure~\ref{fig:ablation2}, DeepRAG achieves lower retrieval costs while maintaining higher average performance. In contrast, the sentence-level partial order pairs learned incorrect preferences, resulting in over-reliance on internal knowledge and consequently leading to both low retrieval costs and poor performance.



%% file: table/efficiency.tex
\begin{table}[t]
  \centering
      \setlength{\abovecaptionskip}{0.5pt} 
  \resizebox{\linewidth}{!}{
  \begin{tabular}{lccccc}
    \toprule
     \multirow{2}[1]{*}{Method} & \multirow{2}[1]{*}{EM} & \multicolumn{3}{c}{Avg. Retrievals} & \multirow{2}[1]{*}{Time} \\
     & & All & Correct & Incorrect & \\
    \midrule
     FLARE        & 28.80  & 0.00  & 0.00  & 0.00  & 2.58  \\
     DRAGIN       & 21.20  & 0.00  & 0.00  & 0.00  & 1.36 \\
     UAR          & 22.70  & 0.96  & 0.95  & 0.97  & 0.43  \\
     TAARE        & 23.40  & 0.66  & 0.65  & 0.66  & 0.11  \\
     IterDRAG     & 15.90  & 2.25  & 2.16  & 2.27  & 1.09 \\
     Auto-RAG     & 17.40  & 4.52  & 3.03  & 2.35  & 0.71  \\
     DeepRAG-Imi  & 30.00  & 0.43  & 0.13  & 0.56  & 0.67 \\
     DeepRAG      & 32.70  & 0.28  & 0.12  & 0.36  & 0.50 \\
    \bottomrule
  \end{tabular}
  }
  \caption{
  Retrieval frequency analysis on WebQuestions across different methods. ``Correct'' indicates the average number of retrievals for instances where the model produced correct answers, while ``Incorrect'' represents the average retrievals for cases with incorrect answers. Time refers to the average seconds spent per item.
  }
  \label{tab:efficiency}
\end{table}

%% file: table/relevance.tex

\begin{table}[t]
  \centering
  \setlength{\abovecaptionskip}{0.5pt} 

  \resizebox{\linewidth}{!}{
    \begin{tabular}{ccccc}
    \toprule
    Method &  F1 &  Acc & Balanced Acc & MCC \\
    \midrule
    FLARE & 0.000  & 0.718  & 0.500  & 0.000 \\
    DRAGIN & 0.007  & 0.709  & 0.495  & -0.045  \\
    UAR   & 0.481  & \textbf{0.756}  & 0.648  & 0.341  \\
    TAARE & 0.127  & 0.712  & 0.518  & 0.078  \\
    Iter-DRAG & 0.000 & 0.718 & 0.500 & 0.000 \\
    Auto-RAG & 0.000 & 0.718 & 0.500 & 0.000  \\
    DeepRAG-Imi & 0.580  & 0.732  & 0.709  & 0.393  \\
    DeepRAG  & \textbf{0.621}  & 0.749  & \textbf{0.743}  & \textbf{0.451} \\
    \bottomrule
    \end{tabular}%
    }
    \caption{Analysis of internal knowledge utilization across different adaptive retrieval methods on 2WikiMultihopQA.}
\label{tab:relevance}%
\end{table}%

%% file: latex/fig_ablation.tex
\begin{figure*}[t]
    \centering
    \setlength{\abovecaptionskip}{0.5pt} 
    \begin{minipage}[b]{0.31\textwidth}
        \centering
        \begin{minipage}[b]{0.98\textwidth}
            \centering
            \resizebox{\textwidth}{!}{
            \begin{tikzpicture}
                \begin{axis}[
                    width=10cm,
                    height=8cm,
                    ybar,
                    xmin=0.5, xmax=3.5,
                    ymin=20, ymax=57,
                    axis y line*=left,
                    xtick={1,2,3},
                    xticklabels={parametric only,DeepRAG,retrieve only},
                    ticklabel style={font=\LARGE},
                    xticklabel style={
                        rotate=15,
                        anchor=center,
                        yshift=-18
                    },
                    label style={font=\LARGE},
                    ylabel={Average Score},
                    ylabel style={font=\huge},
                    bar width=0.4,
                    legend style={
                        anchor=north east,       
                        font=\normalsize,
                        cells={anchor=west},
                    },
                ]
        
                \addplot[
                    fill={rgb,255:red,123; green,167; blue,157},
                    nodes near coords,
                    point meta=y,
                    every node near coord/.append style={font=\Large, yshift=2pt}
                ] coordinates {
                    (0.8,25.32)
                    (1.8,47.67)
                    (2.8,45.39)
                };
                \addlegendentry{Avg. Score}
            
                \addlegendimage{fill={rgb,255:red,244; green,177; blue,131}}
                \addlegendentry{Avg. Retrievals}
            
                \end{axis}
        
                \begin{axis}[
                    width=10cm,
                    height=8cm,
                    ybar,
                    xmin=0.5, xmax=3.5,
                    ymin=0, ymax=4.2,
                    axis x line=none,
                    axis y line*=right,
                    ticklabel style={font=\LARGE},
                    label style={font=\LARGE},
                    ylabel={Average Retrievals},
                    ylabel style={font=\huge},
                    bar width=0.4,
                ]
            
                \addplot[
                    fill={rgb,255:red,244; green,177; blue,131},
                    nodes near coords,
                    point meta=y,
                    every node near coord/.append style={font=\Large, yshift=2pt}
                ] coordinates {
                    (1.2,0.0)
                    (2.2,1.368912773)
                    (3.2,2.768780685)
                };
            
                \end{axis}
            
            \end{tikzpicture}
            }
        \end{minipage}
        \caption{Comparative analysis of retrieval strategies: parametric only or retrieve only.}
        \label{fig:analysis1}
    \end{minipage}
    \hfill
    \begin{minipage}[b]{0.65\textwidth}
        \centering
        \begin{minipage}[b]{0.48\textwidth}
            \centering
            \resizebox{\textwidth}{!}{
            \begin{tikzpicture}
                \begin{axis}[
                    width=10cm,
                    height=8cm,
                    ybar,
                    xmin=0.5, xmax=3.5,
                    ymin=40, ymax=46,
                    axis y line*=left,
                    xtick={1,2,3},
                    xticklabels={DeepRAG-Imi,most,random},
                    ticklabel style={font=\LARGE},
                    xticklabel style={
                        rotate=15,
                        anchor=center,     
                        yshift=-20
                    },
                    label style={font=\LARGE},
                    ylabel={Average Score},
                    ylabel style={font=\huge},
                    bar width=0.4,
                    legend style={
                        anchor=north east,       
                        font=\normalsize,
                        cells={anchor=west},
                    }
                ]

                \addplot[
                    fill={rgb,255:red,123; green,167; blue,157},
                    nodes near coords,
                    point meta=y,
                    every node near coord/.append style={font=\Large, yshift=2pt}
                ] coordinates {
                    (0.8,44.60)
                    (1.8,41.12)
                    (2.8,40.56)
                };
                \addlegendentry{Avg. Score}
            
                \addlegendimage{fill={rgb,255:red,244; green,177; blue,131}}
                \addlegendentry{Avg. Retrievals}
            
                \end{axis}
        
                \begin{axis}[
                    width=10cm,
                    height=8cm,
                    ybar,
                    xmin=0.5, xmax=3.5,
                    ymin=0.5, ymax=3,
                    axis x line=none,
                    axis y line*=right,
                    ticklabel style={font=\LARGE},
                    label style={font=\LARGE},
                    ylabel={Average Retrievals},
                    ylabel style={font=\huge},
                    bar width=0.4,
                ]
            
                \addplot[
                    fill={rgb,255:red,244; green,177; blue,131},
                    nodes near coords,
                    point meta=y,
                    every node near coord/.append style={font=\Large, yshift=2pt}
                ] coordinates {
                    (1.2,0.98)
                    (2.2,2.11)
                    (3.2,1.81)
                };
    
            \end{axis}
        \end{tikzpicture}
    }
        \caption{Average score and retrievals on the ablation study for Imitation Learning.}
        \label{fig:ablation1}
    \end{minipage}
        \hfill
        \begin{minipage}[b]{0.485\textwidth}
            \centering
            \resizebox{\textwidth}{!}{
            \begin{tikzpicture}
                \begin{axis}[
                    width=10cm,
                    height=8cm,
                    ybar,
                    xmin=0.5, xmax=3.5,
                    ymin=20, ymax=55,
                    axis y line*=left,
                    xtick={1,2,3},
                    xticklabels={DeepRAG,all-node,sentence-wise},
                    ticklabel style={font=\LARGE},
                    xticklabel style={
                        rotate=15,
                        anchor=center,     
                        yshift=-20
                    },
                    label style={font=\LARGE},
                    ylabel={Average Score},
                    ylabel style={font=\huge},
                    bar width=0.4,
                    legend style={
                        anchor=north east,       
                        font=\normalsize,
                        cells={anchor=west},
                    }
                ]

                \addplot[
                    fill={rgb,255:red,123; green,167; blue,157},
                    nodes near coords,
                    point meta=y,
                    every node near coord/.append style={font=\Large, yshift=2pt}
                ] coordinates {
                    (0.8,47.67)
                    (1.8,45.30)
                    (2.8,21.14)
                };
                \addlegendentry{Avg. Score}
            
                \addlegendimage{fill={rgb,255:red,244; green,177; blue,131}}
                \addlegendentry{Avg. Retrievals}
            
                \end{axis}
        
                \begin{axis}[
                    width=10cm,
                    height=8cm,
                    ybar,
                    xmin=0.5, xmax=3.5,
                    ymin=0.5, ymax=2.0,
                    axis x line=none,
                    axis y line*=right,
                    ticklabel style={font=\LARGE},
                    label style={font=\LARGE},
                    ylabel={Average Retrievals},
                    ylabel style={font=\huge},
                    bar width=0.4,
                ]
            
                \addplot[
                    fill={rgb,255:red,244; green,177; blue,131},
                    nodes near coords,
                    point meta=y,
                    every node near coord/.append style={font=\Large, yshift=2pt}
                ] coordinates {
                     (1.2,1.37)
                    (2.2,1.56)
                    (3.2,0.57)
                };
    
            \end{axis}
        \end{tikzpicture}
    }
        \caption{Average score and retrievals on the ablation study for Chain of Calibration. }
        \label{fig:ablation2}
    \end{minipage}
    \end{minipage}
\end{figure*}

%% file: figure/pie.tex
\begin{figure}[t]
    \centering
    \begin{minipage}[]{0.21\textwidth}
        \centering
        \scriptsize
        {\begin{tikzpicture}[scale=1.0,baseline=(current bounding box.center)]
            \pie[text=legend, 
                sum=1000, 
                radius=1,
                rotate=-97,
                color={soft1, soft2, soft3, soft4, soft5}]{
                683/3, 
                65/4, 
                61/5, 
                191/$\geq$6
            }
        \end{tikzpicture}}
        \vspace{-0.2cm}
        \hspace{-1.1cm}\normalsize (a)
    \end{minipage}
    \hfill
    \begin{minipage}[]{0.21\textwidth}
        \centering
        \scriptsize
        {\begin{tikzpicture}[scale=1.0,baseline=(current bounding box.center)]
            \pie[text=legend, 
                sum=1000, 
                radius=1,
                rotate=310,
                color={soft1, soft2, soft3, soft4, soft5}]{
                381/0, 
                177/1, 
                433/2, 
                9/$\geq$3
            }
        \end{tikzpicture}}
        \hspace{-1.cm}\normalsize (b)
    \end{minipage}    
    \caption{\hspace{-0.2em}(a) Subquery Statistics. (b) Retrieval Statistics.}
    \label{pie}
\end{figure}

%% file: table/word_count.tex
\begin{figure}[t]
    \centering
    \setlength{\abovecaptionskip}{0.5pt} 
    \begin{minipage}{\textwidth}
    \hspace{2em} 
    \begin{minipage}[b]{0.31\textwidth}
        \centering
        \resizebox{\textwidth}{!}{
            \begin{tikzpicture}
                \begin{axis}[
                    ybar=0pt,
                    width=12cm,
                    height=8cm,
                    bar width=25pt,
                    symbolic x coords={WH,Noun,Verb,And/Or},
                    xtick=data,
                    ymin=0,
                    ymax=2.3,
                    yticklabel style={font=\Large},
                    label style={font=\LARGE},
                    ylabel={Average Count},
                    ylabel style={font=\LARGE},
                    legend pos=north west,
                    nodes near coords style={/pgf/number format/.cd,fixed,fixed zerofill,precision=2},
                    nodes near coords,
                    enlarge x limits=0.15,
                    x tick label style={font=\Large},
                ]

                \addplot[
                    fill={rgb,255:red,123; green,167; blue,157},
                    bar shift=-12.5pt  
                ] coordinates {
                    (WH,0.98) (Noun,0.04) (Verb,1.13) (And/Or,0.12)
                };
                \addlegendentry{DeepRAG}

                \addplot[
                    fill={rgb,255:red,244; green,177; blue,131},
                    bar shift=12.5pt  
                ] coordinates {
                    (WH,1.26) (Noun,0.30) (Verb,2.05) (And/Or,0.77)
                };
                \addlegendentry{Auto-RAG}
                \end{axis}
            \end{tikzpicture}
        }
    \end{minipage}
    \end{minipage}
    \caption{Average counts of WH-words, nouns, verbs, and conjunctions (and/or) per subquery.}
    \label{fig:atomic}
\end{figure}

%% file: latex/conclude.tex
\vspace{-0.15cm}
\section{Conclusion}
\vspace{-0.15cm}

In this paper, we present DeepRAG to model retrieval-augmented reasoning as a Markov Decision Process, enabling strategic and adaptive retrieval by decomposing queries into subqueries and retrieval on demand.
Specifically, we develop a binary tree search method to synthesize data for imitation learning and further chain of calibration to train the model in an end-to-end manner.
Experiments across various QA tasks show that DeepRAG improves retrieval efficiency while improving answer\hspace{-0.1em} accuracy \hspace{-0.1em}by 26.4\%, \hspace{-0.2em}demonstrating its effectiveness in optimizing retrieval-augmented reasoning.

%% file: latex/limitation.tex
\section*{Limitations}
There are several limitations of our current DeepRAG framework, which we plan to address in the future. 
Firstly, we construct datasets based on the final answer accuracy using exact match score.
In the future, we will expand our work to more domains like multi-turn dialogue with richer metrics like perplexity. 
Secondly, despite our method showing strong generalization across multi-hop factual QA, time-sensitive QA, and heterogeneous knowledge base QA, it lacks integration with external resources such as knowledge graphs and tools. 
 We will expand our work to domains requiring diverse external information integration, including retrieved data, knowledge graph data, and tool output.

%% file: latex/appendix.tex
\newpage
\appendix
\section{Templates}

\begin{figure*}[t]
    \centering
    \includegraphics[width=1.0\linewidth]{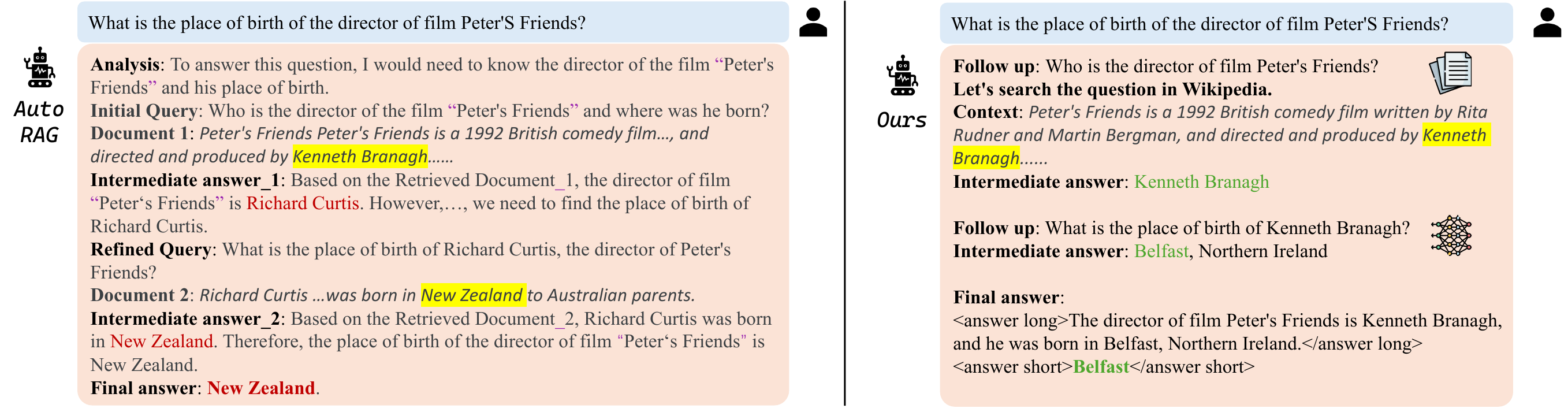}
    \caption{Case Study: Auto-RAG vs. DeepRAG. DeepRAG achieves success by atomic query decomposition, faithful intermediate answer, and adaptively using internal knowledge.}
    \label{fig:case-study}
\end{figure*}
\subsection{Case Study}

\subsection{DeepRAG Construct Instruction}
\label{template}
\begin{tcolorbox}
\small

Instruction: You are a helpful Retrieval-Augmented Generation (RAG) model. Your task is to answer questions by logically decomposing them into clear sub-questions and iteratively addressing each one. \\

Use "Follow up:" to introduce each sub-question and "Intermediate answer:" to provide answers. \\

For each sub-question, decide whether you can provide a direct answer or if additional information is required. If additional information is needed, state, "Let's search the question in Wikipedia." and then use the retrieved information to respond comprehensively. If a direct answer is possible, provide it immediately without searching.

\end{tcolorbox}


\section{Method Details}
\subsection{Imitation Learning Objective}
\label{eq1}
We implement a masked loss function for the retrieved documents to prevent the model from learning irrelevant or noisy text that could negatively impact its performance. In this way, we hope the model to enhance the ability to decompose subqueries and retrieve them based on demand. 
For each instance, the loss function is formulated as follows:

\begin{small}
    
\begin{align}
&\mathcal{L}= -\sum_{1\leq i\leq n}\text{log }[\text{Pr}(q_i| s_{i-1}) + \text{Pr}(a_i| s_{i-1}, q_i, d_i)] \notag
\end{align}
\end{small}
where, $d_i$ refers to \textit{null} if there is no reieval for $i$th reasoning step, $n$ refers to the total iteration.
\subsection{Chain of Calibration Objective}
\label{eq2}
We fine-tune the LLM using a Chain of Calibration objective on our synthesized preference data. 

Given the $i$-th subquery and the state $s_i = [x,q_1, r_1, \cdots, q_{i-1}, r_{i-1}]$, we have two distince intermediate answer $r_i^1 = a_i^1$ and $r_i^2 = (d_i, a_i^2)$. Based on the process above, we have known which $r_i$ is preferred.  As a result, the training objective can be formulated as follows:

\begin{small}
\begin{equation}
\mathcal{L} = 
-  \log \sigma \left( \beta \log \frac{\pi_{\theta}(y_w \mid s_i, q_i)}{\pi_{\text{ref}}(y_w \mid s_i, q_i)} - \beta \log \frac{\pi_{\theta}(y_l \mid s_i, q_i)}{\pi_{\text{ref}}(y_l \mid s_i, q_i)} \right)  \notag
\end{equation}
\end{small}
where $\sigma$ is the logistic function, the hyperparameter $\beta$ regulates the penalty imposed for the deviations
from the base reference model $\pi_{ref}$.
The terms \( y_w \) and \( y_l \) refer to the generated snippets for direct answers and retrieved answers, respectively. Specifically, the snippet ``Intermediate Answer:'' corresponds to a direct answer, while the snippet ``Let's search the question on Wikipedia'' corresponds to retrieval-based answers.

\section{Detailed Analysis}

As illustrated in Figure~\ref{fig:case-study}, we conduct a case study comparing DeepRAG with Auto-RAG~\cite{yu2024autorag}, a closely related method that utilizes iterative retrieval for retrieval-augmented generation.
For each subquery, Auto-RAG retrieves relevant documents and generates a corresponding subanswer. This approach is not only time-consuming but also fails when no relevant documents are retrieved.
Although Auto-RAG attempts to address this issue using its own relevant documents, it falls into endless loops in most cases.
In contrast, DeepRAG iteratively generates subqueries and determines whether to use internal knowledge at each iteration. The binary tree search data synthesis method for optimization ensures reliable subquery generation, intermediate answers, and final answers. Even when no related information exists in retrieved documents, the model is directed to provide a final answer based on internal knowledge.

\subsection{Retrieval Efficiency}

To demonstrate the efficiency of our method, we compare the average number of retrievals on 2WikiMultihopQA and WebQuestions. As shown in Table~\ref{tab:efficiency}, We have following observations:

1) Compared to other adaptive retrieval methods, DeepRAG can achieve higher accuracy with relatively lower retrieval costs. This can be attributed to our dynamic usage of internal knowledge.
Additionally, DeepRAG exhibits a positive trend in exploring relevant evidence when faced with insufficient retrieval results, as evidenced by the lower average retrieval numbers in both 2WMQA (0.92 compared to 1.25) and WQ (0.12 compared to 0.36).
2) Confidence-based approaches demonstrate limited robustness across datasets.  
For instance, while using identical thresholds, both FLARE and DRAGIN methods show inconsistent behaviors: they trigger approximately one retrieval per query in 2WMQA, but fail to reach the retrieval threshold entirely in WQ. This inconsistency highlights the challenge of maintaining reliable performance across different datasets using confidence-based methods.
3) Iterative retrieval-based approaches typically require numerous retrieval operations, resulting in substantial computational costs. Therefore, efficient adaptive retrieval methods like DeepRAG become crucial for optimizing resource utilization while maintaining performance.

\subsection{Relevance to Parametric Knowledge}
\label{relevance}
In this section, we investigate the relationship between retrieval needs and parametric knowledge to demonstrate how effectively our method explores the knowledge boundary.

Ideally, models should initiate retrieval for queries beyond their parametric knowledge while utilizing their existing knowledge for familiar queries. We use CoT results as an indicator of whether the model can answer questions using its parametric knowledge. Subsequently, we analyze whether other adaptive retrieval methods align with this pattern of parametric knowledge utilization.
We evaluate the relevance using four metrics. F1 score and Accuracy serve as basic performance measures, while balanced accuracy and Matthews Correlation Coefficient(MCC) are employed to account for the class imbalance between retrieval-required and retrieval-not-required cases. The MCC ranges from -1 to 1, where a value of 1 indicates perfect correlation, 0 represents no correlation (random chance), and -1 signifies an inverse correlation.

As shown in Table~\ref{tab:relevance}, we find that 1) 
DeepRAG demonstrates superior relevance performance across F1, balanced accuracy, and MCC metrics. 
This suggests that DeepRAG successfully identifies retrieval necessity by exploring knowledge boundary.
2) While FLARE, DRAGIN, and TAARE exhibit high accuracy scores, their relatively low balanced accuracy and MCC scores suggest they mainly succeed in retrieval-required cases but struggle to properly avoid unnecessary retrievals.

\subsection{Performance against Strong Baseline Models}


We compare DeepRAG with recent strong reasoning models: QwQ-32B-preview~\cite{qwq-32b-preview} and gpt-4o-turbo~\cite{OpenAI_hello_gpt4o}. 
As shown in Table~\ref{tab:strong}, DeepRAG achieves superior average performance over QwQ and gpt-4o, particularly in time-sensitive QA tasks. 
While DeepRAG does not surpass gpt-4o in some cases, it achieves comparable performance levels. 
These results demonstrate that by adaptively leveraging retrieval, DeepRAG can achieve an equivalent level of factual accuracy to the parametric knowledge of strong reasoning models.

\input{table/strong_model}

\subsection{Ablation Study}
In this section, we conducted experiments to validate the effectiveness of DeepRAG's data construction and training process. 

%

\paragraph{Imitation Learning}

\begin{table}[htbp]
  \centering
  \resizebox{\linewidth}{!}{
    \begin{tabular}{rccccc}
    \toprule
    Method & ID & CAG   & PopQA & WebQuestion & \multirow{2}[2]{*}{Avg} \\
             & F1    & EM    & EM    & EM    & \\
    \midrule
    \multicolumn{1}{l}{DeepRAG-Imi} &  \textbf{49.46}  & 50.47  & \textbf{43.60 } & \textbf{30.00 } & \textbf{44.60 } \\
    most & 47.31   & 51.09  & 31.30  & 28.00  & 41.12  \\
    random & 44.76   & \textbf{51.40 } & 34.80  & 27.10  & 40.56  \\
    \bottomrule
    \end{tabular}%
    }
    \caption{Experiment results of the ablation study on the Imitation Learning Stage. ID refers to the average score of two in-distribution dataset HotpotQA and 2WikiMultihopQA.}
  \label{tab:sft-abla}%
\end{table}%

We compare our default strategy of selecting paths with minimal retrieval cost against two alternative approaches: maximum retrieval cost and random path selection.
As shown in Table~\ref{tab:sft-abla}, DeepRAG-Imi enables the model to learn knowledge boundaries during the imitation learning stage. 
Notably, CAG performs relatively poorly at this stage due to its time-sensitive nature, which necessitates constant retrieval of up-to-date information. 
Moreover, as illustrated in Figure~\ref{fig:ablation1}, DeepRAG-Imi achieves lower retrieval costs and higher average performance compared to both the maximum-retrieval-cost and random selection methods.


\input{table/abaltion}

\paragraph{Chain of Calibration} We compare our default approach of constructing preferences based on nodes from optimal paths against two alternatives: constructing pairs for all nodes and constructing sentence-level partial order pairs based on retrieval efficiency.
As shown in Table~\ref{tab:dpo-abla}, DeepRAG demonstrates significant advantages over both variants. Specifically, as illustrated in Figure~\ref{fig:ablation2}, DeepRAG achieves lower retrieval costs while maintaining higher average performance. In contrast, the sentence-level partial order pairs learned incorrect preferences, resulting in over-reliance on internal knowledge and consequently leading to both low retrieval costs and poor performance.

\subsection{Implementation Details under RL Setting}
We implement based on Search-R1 repository \footnote{\url{https://github.com/PeterGriffinJin/Search-R1}}. 
We adopt GRPO with a batch size of 32 and perform 8 rollouts per prompt. To avoid introducing noise, we additionally mask the retrieved text during training.

\label{appendix:rl}

\begin{equation}
R_t = \left\{
\begin{array}{ll}
0,  \text{answer \ding{55} and format \ding{55}} \\
0.1,  \text{answer \ding{55} and format \ding{51}} \\
1 - 0.1 \times \min(5, \text{retrieve\_time}_t),  \text{answer \ding{51}}
\end{array}
\right.
\notag
\end{equation}

\subsection{Retrieval Efficiency}
\label{appendix:efficiency}




\input{table/appendix_efficiency}
To demonstrate the efficiency of our method, we compare the average number of retrievals on 2WikiMultihopQA and WebQuestions. As shown in Table~\ref{tab:efficiency_appendix}, We have the following observations:
1) DeepRAG can achieve higher accuracy with relatively lower retrieval costs, attributed to its dynamic usage of internal knowledge.
2) Confidence-based approaches demonstrate limited robustness across datasets.  
For instance, neither FLARE nor DRAGIN trigger retrieval under the default confidence threshold in WQ.
%
3) Iterative retrieval-based methods typically require numerous retrieval operations.
Therefore, efficient adaptive retrieval methods like DeepRAG become crucial for optimizing resource utilization while maintaining performance.

%% file: table/strong_model.tex

\begin{table}[htbp]
  \centering
  \resizebox{\linewidth}{!}{
    \begin{tabular}{cccccc}
    \toprule
    Models & ID & CAG   & PopQA & WQ    & Avg \\
    \midrule
    QwQ-32B & 31.43  & 3.43  & 10.60  & 15.10  & 18.40  \\
    gpt-4o-turbo & \textbf{60.6}  & 23.36  & 43.50  & 25.35  & 42.68  \\
    DeepRAG-qwen & 43.00  & 51.09  & 40.60  & 24.20  & 40.38  \\
    DeepRAG-llama & 52.40  & \textbf{52.96}  & \textbf{42.50}  & \textbf{32.70}  & \textbf{46.59}  \\
    \bottomrule
    \end{tabular}%
    }
    \caption{Performance against strong baseline models.}
    \vspace{-0.2cm}
  \label{tab:strong}%
\end{table}%

%% file: table/abaltion.tex
\begin{table}[t]
  \centering
  \resizebox{\linewidth}{!}{
    \begin{tabular}{rccccc}
    \toprule
    Method & ID & CAG   & PopQA & WebQuestion & \multirow{2}[2]{*}{Avg}\\
        & F1    & EM    & EM    & EM    &  \\
    \midrule
    \multicolumn{1}{l}{DeepRAG} & \textbf{52.40} & \textbf{61.92 } & \textbf{47.80 } & \textbf{45.24 } & \textbf{47.67 } \\
    all-node & 50.92   & 50.47  & 41.50  & 32.70  & 45.30  \\
    sentence-wise & 30.16   & 12.46  & 20.00  & 12.90  & 21.14 \\
    \bottomrule
    \end{tabular}%
    }
  \caption{Experiment results of the ablation study on the Chain of Calibration Stage.}
  \label{tab:dpo-abla}%
\end{table}%

%% file: table/appendix_efficiency.tex
\begin{table}[t]
  \centering
  \resizebox{\linewidth}{!}{
\begin{tabular}{llcccc}
\toprule
\multirow{2}[1]{*}{Dataset}  &  \multirow{2}[1]{*}{Method}   & \multirow{2}[1]{*}{EM} & \multicolumn{3}{c}{Avg. Retrievals} \\
      &       &       & All   & Correct & Incorrect \\
\midrule
\multirow{7}[2]{*}{2WMQA} & FLARE & 30.30  & 0.99  & 1.00  & 0.99  \\
      & DRAGIN & 29.10  & 1.03  & 1.03  & 1.03  \\
      & UAR & 34.80 & 0.81  & 0.68 &   0.89\\
      & TAARE & 35.20  & 0.93  & 0.93  & 0.97  \\
      & IterDRAG & 19.60 & 2.46 & 2.49 &  2.45 \\
      & Auto-RAG & 23.00  & 6.26  & 4.13  & 1.81  \\
      & DeepRAG-Imi & 47.20  & 1.13  & 0.95  & 1.28  \\
      & DeepRAG  & 48.10  & 1.09  & 0.92  & 1.25  \\
\hline
\multirow{7}[2]{*}{WQ} & FLARE & 28.80  & 0.00  & 0.00  & 0.00 \\
      & DRAGIN & 21.20  & 0.00  & 0.00  & 0.00  \\
      & UAR & 22.70 & 0.96 & 0.95 & 0.97\\
      & TAARE & 23.40  & 0.66  & 0.65  & 0.66  \\
      & IterDRAG & 15.90 & 2.25 & 2.16 & 2.27 \\
      & Auto-RAG & 17.40  & 4.52  & 3.03  & 2.35  \\
      & DeepRAG-Imi & 30.00  & 0.43  & 0.13  & 0.56  \\
      & DeepRAG  & 32.70  & 0.28  & 0.12  & 0.36  \\
\bottomrule
\end{tabular}%
}
\caption{
Retrieval frequency analysis on 2WikiMultihopQA(2WMQA) and WebQuestions(WQ) across different adaptive retrieval methods. "Correct" indicates the average number of retrievals for instances where the model produced correct answers, while "Incorrect" represents the average retrievals for cases with incorrect answers.
}
\label{tab:efficiency_appendix}%
\end{table}%